\documentclass[10pt,twocolumn,letterpaper]{article}

\usepackage{ijcb}
\usepackage{times}
\usepackage{epsfig}
\usepackage{graphicx}
\usepackage{amsmath}
\usepackage{amssymb}
\usepackage[norule,symbol,perpage]{footmisc}

\usepackage{multirow}
\usepackage{subfigure}
\usepackage{array}
\usepackage{threeparttable}

\usepackage{hyperref}
\hypersetup{colorlinks,allcolors=black}

\ijcbfinalcopy 

\ifijcbfinal\fi

\makeatletter
\def\ps@IEEEtitlepagestyle{
\def\@oddfoot{\mycopyrightnotice}
\def\@evenfoot{}
}
\def\mycopyrightnotice{
{\hfill \footnotesize 978-1-7281-9186-7/20/\$31.00 \copyright 2020 IEEE\hfill}
}
\makeatother

\begin{document}

\title{Dense-View GEIs Set: View Space Covering for Gait Recognition \\ based on Dense-View GAN}

\author{Rijun Liao$^1$, Weizhi An$^2$, Shiqi Yu$^3$, Zhu Li$^1$, Yongzhen Huang$^{4,5}$\\
 \and \and
 $^1${\normalsize Department of Computer Science and Electrical
Engineering,University of Missouri-Kansas City, USA.} \\ 
 $^2${\normalsize Department of Computer Science and Engineering, University of Texas at Arlington, USA} \\
 $^3${\normalsize Department of Computer Science and Engineering, Southern University of Science  and Technology, China} \\
 $^4${\normalsize  National Laboratory of Pattern Recognition, Institute of Automation,
Chinese Academy of Sciences, China} \\
 $^5${\normalsize Watrix technology limited co. ltd, China}
 \and \and
 {\tt\small rijun.liao@mail.umkc.edu, weizhi.an@mavs.uta.edu, yusq@sustech.edu.cn,} \\ {\tt\small lizhu@umkc.edu, yongzhen.huang@nlpr.ia.ac.cn}}

\maketitle
\thispagestyle{empty}

\begin{abstract}
        Gait recognition has proven to be effective for long-distance human recognition. But view variance of gait features would change human appearance greatly and reduce its performance. Most existing gait datasets usually collect data with a dozen different angles, or even more few. Limited view angles would prevent learning better view invariant feature. It can further improve robustness of gait recognition if we collect data with various angles at $1^\circ$ interval. But it is time consuming and labor consuming to collect this kind of dataset. In this paper, we, therefore, introduce a \textit{Dense-View GEIs Set (DV-GEIs)} to deal with the challenge of limited view angles. This set can cover the whole view space, view angle from $0^\circ$ to $180^\circ$ with $1^\circ$ interval. In addition, \textit{Dense-View GAN (DV-GAN)} is proposed to synthesize this dense view set. \textit{DV-GAN} consists of \textit{Generator},  \textit{Discriminator} and \textit{Monitor}, where \textit{Monitor} is designed to preserve human identification and view information. The proposed method is evaluated on the CASIA-B and OU-ISIR dataset. The experimental results show that DV-GEIs synthesized  by \textit{DV-GAN} is an effective way to learn better view invariant feature. We believe the idea of dense view generated samples will further improve the development of gait recognition.
\end{abstract}

\let\thefootnote\relax\footnotetext{\mycopyrightnotice}

\section{Introduction}

Gait is one kind of popular biometric features for human identification. Compared with other features like face, iris, palmprint and fingerprint, gait provides a unique possibility to identify a subject at a long distance without people's cooperation. Therefore, it has great potential application in catching criminals,  video surveillance and social security.

However, gait recognition is still challenging in the real application. This is because there are many variations would reduce its performance. View is one of gait challenges because we can not control the walking direction of people and view changing that lead the human body shape to change greatly. In this paper, we propose DV-GEIs to further reduce the influence of view variation.

	\begin{figure}
		\centering
		\includegraphics[height=2.4cm]{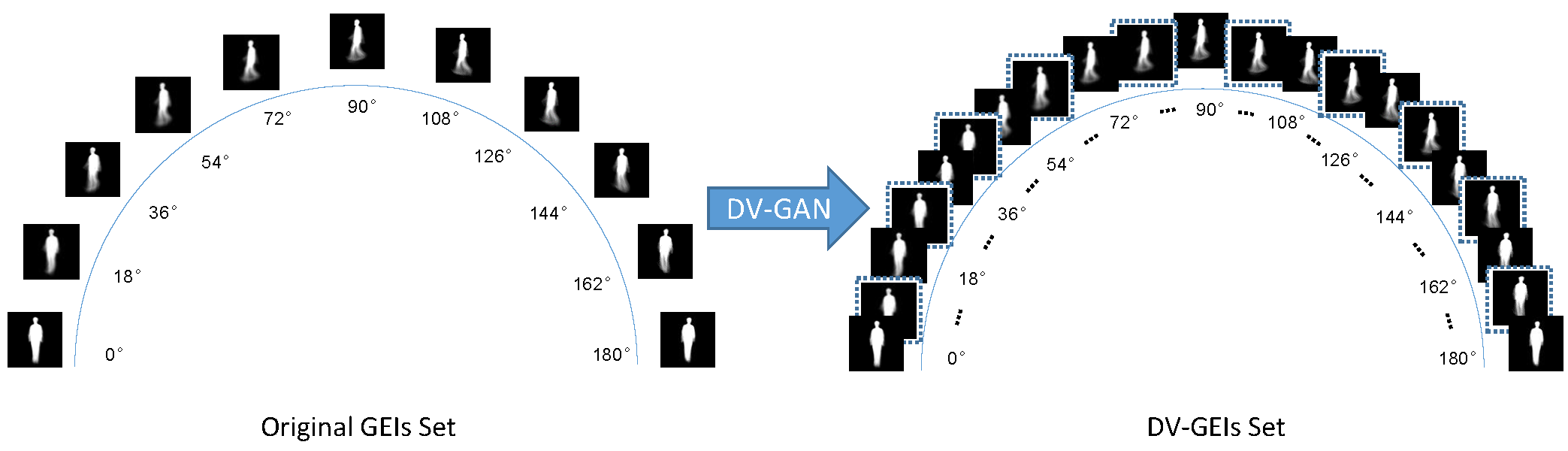}
		\caption{
		Dense-View GEIs Set (DV-GEIs): view space covering for learning better view invariant feature, view angle range from $0^\circ$ to $180^\circ$ with $1^\circ$ interval. DV-GAN is proposed to synthesize realistic samples with various view angle conditions. (Sample images from CASIA-B dataset~\cite{yu2006framework})}
		\label{fig:framework}
	\end{figure}
	
In order to solve the cross-view challenge, some researchers focus on view transformation model (VTM) which can transform gait features from one view to another view. Examples are  FD-VTM~\cite{Makihara2006Gait},  RSVD-VTM~\cite{Kusakunniran2010Multiple} and RPCA-VTM~\cite{Zheng2011Robust}. But most VTM methods need to know angles of probe and gallery before extracting gait features. This means that each view needs one model which has some challenges in the real application. In the next few years, SPAE~\cite{Yu2017View}, GaitGAN~\cite{Yu2017GaitGAN} and GaitGANv2~\cite{yu2019gaitganv2} are proposed that can transform any view gait into the side view gait by using only an uniform model.  However, the side view transform strategy will collapse when the view variance is large. 

With the development of deep learning, some works~\cite{liao2017pose,an2018pose,liao2020model,an2020performance} only use several human pose coordinates as input gait features which is robust to human appearance. However, its performance still need to be improved because human pose coordinates have not enough information compared with human appearance. In addition, Wu {\it et al.}~\cite{Wu2017A} and Chao {\it et al.}~\cite{chao2019gaitset} directly take a sequence of human silhouettes as input data rather than using the hard-crafted gait features and achieve high performance. The price of these methods is high computational cost. 

The above methods have made a great contribution to the development of gait recognition. We think that it can further improve its robustness to view variation if we can collect data with more view angle. Because most existing datasets not cover all kinds of view condition, and deep learning depends heavily on big data. CASIA-B dataset~\cite{yu2006framework} has 11 views with $18^\circ$ view angle interval, OU-MVPL~\cite{takemura2018multi} has 14 views and OU-ISIR~\cite{Iwama_IFS2012} only has four views. It has negative influence to learn better invariant feature if the number of view angle in the dataset is limited. We can  collect data with more view angles manually. But it is time consuming and labor consuming to collect dense view angle data.

Recently, some researchers use synthesized samples based on GAN to improve original performance. Chen {\it et al.}~\cite{chen2018image} used GAN to generate noise samples and use them in image blind denoising. Qian {\it et al.}~\cite{qian2018pose} combined human pose and GAN to synthesize human image in specific pose for person re-identification. Those methods can greatly improve its original performance. But this idea has not been achieved in gait recognition task, because it needs to find a suitable solution to synthesize samples with different conditions. One popular based on GAN work, GaitGAN~\cite{Yu2017GaitGAN}, has used GAN to transform any view GEI into the side view GEI, which effectively improve gait robustness. In our work, we extend the development of GAN in gait recognition by generating samples with more views, rather than view transformation.

Unlike above view transformation methods, we generate gait features with dense views to improve recognition rate on the cross-view condition.  We are inspired by the idea of synthesized samples and GaitGAN~\cite{Yu2017GaitGAN}, and proposed a novel generation model DV-GAN to synthesize DV-GEIs set.  GEI~\cite{Han2006Individual} is employed to be the gait features in the proposed method same as GaitGAN~\cite{Yu2017GaitGAN} and SPAE~\cite{Yu2017View}, because its robustness to noise and its efficiency in computation. Our method in this paper has the following contributions: 

	\begin{itemize}
		\item We introduce a novel Dense-View GEIs Set (DV-GEIs) to  solve the challenge of limited view angles on the existing gait dataset.  Most existing datasets usually capture samples with several or dozens view  at large interval, while the view angle of DV-GEIs could cover whole view space to make up for small number of view angles, range from $0^\circ$ to $180^\circ$ with $1^\circ$ interval, as shown in Figure~\ref{fig:framework}. 
		
		\item A novel GEI generation model Dense-View GAN (DV-GAN) is proposed to generate realistic GEIs with various view angle conditions. Compared with traditional GAN~\cite{Goodfellow2014GAN} which has generator and discriminator, DV-GAN includes additional monitor which can maintain human identification and view information very well. 
		
		\item We have performed several experiments on CASIA-B~\cite{yu2006framework} and OU-ISIR~\cite{Iwama_IFS2012} dataset.  Experimental results shows that dense view samples synthesized by DV-GAN can further improve robustness to view variation compared with original dataset.
		
	\end{itemize}

	\section{Method}

	\subsection{Dense-View GEIs Set (DV-GEIs)}
	\label{GEI synthesizing}

\begin{figure}[htbp]
		\centering
		\includegraphics[height=6cm]{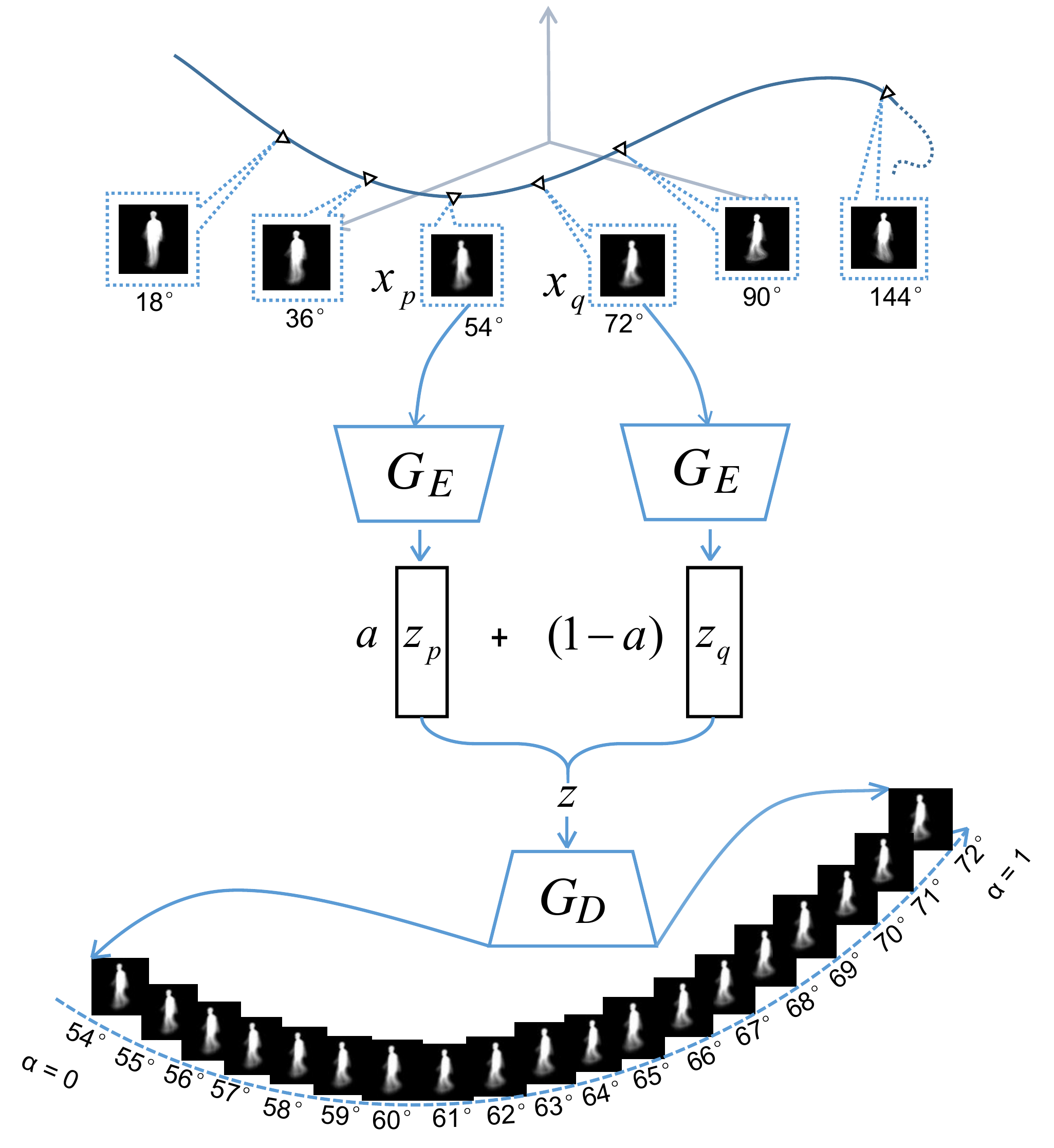}
		\caption{The latent space $z_p$, $z_q$ are encoded by encoder $G_E$ with two input GEIs $x_p$, $x_q$. We synthesize various views GEIs from $p$ angle to $q$ angle by decoding the latent space $z$, where  $z = \alpha z_p + (1 -\alpha){z_q}$,  $\alpha\in[0,1]$. }
		\label{fig:manifold}
	\end{figure}

	We denote the views of GEI as $p, q$ which are correspondent with the input GEI $x_p, x_q$. Our aim is to synthesize various views GEIs, from $p$ angle to $q$ angle, as shown in Figure~\ref{fig:manifold}.  The synthesized GEIs are created according to the following equation:
		\begin{equation}\label{synthesize}
		x'=\{G_D(z) | z = \alpha z_p + (1-\alpha){z_q} \}
		\end{equation}
		
	where $z_p = G_E (x_p),z_q = G_E (x_q)$, latent space $z_p,z_q$ are encoded by encoder $G_E$ which keep the characteristic of gait attribute. The interpolation is defined by linear transformation $z = \alpha z_p + (1-\alpha){z_q}$, where $\alpha\in[0,1]$, and then $z$ is fed to decoder $G_D$  to generate new GEIs.

	The idea of DV-GEIs is inspired by the idea of Hou {\it et al.}~\cite{hou2017deep} which can generate a series of different view angle human faces from \textit{left face} to \textit{right face} by linear transformation $z = \alpha z_p + (1-\alpha){z_q}$ in latent space. In~\cite{hou2017deep},
	authors model the face attribute distribution by training lots of faces with autoencoder, and 	produce latent vectors that can capture the semantic information of face expressions. In addition, Hou {\it et al.}~\cite{hou2017deep} investigate the latent space and show that semantic relationship between different latent representations can be used in facial attribute prediction. In our work, we take advantage of latent space to deal with the challenge of samples with limited view angle in gait recognition task.

	\subsection{Dense-View GAN (DV-GAN)}
	\label{DV-GAN}
	We propose DV-GAN to generate realistic GEIs at various view angle conditions. With the rapid development of adversarial generative network (GAN)~\cite{Goodfellow2014GAN,qian2018pose}, GAN can generate images with realistic details. Our DV-GAN model consists of three neural networks: generator $G$, monitor $M$ and discriminator $D$   as shown in Figure~\ref{fig:GAN}.
	
	\begin{itemize}
		\item \textbf{Generator}:  Given an input GEI $x$, and a target GEI $\hat{x}$, where $x=\hat{x}$. The purpose of our generator is to reconstruct GEI and model gait attribute distribution in latent space. We borrow the pixels to pixels level idea~\cite{pix2pix2017} to reconstruct the GEI image, that is adding $L_1$  norm loss to ensure the output GEI $\hat{x}=G (z,x)$ is the same as input $x$, our generator loss is defined as:
		\begin{equation}\label{L1 funtion}
		\mathop{\min}_{E,G}\mathop{L_{L1} (G (z),x)}
		\end{equation}
		
		U-Net structure (add skip connections between two layers) is employed in our generator, as U-Net~\cite{ronneberger2015u} architecture allows low-level information to shortcut across the network and effectively improve the quality of the generated images. We divide U-Net into two parts, encoder $G_E\{e_1\}$ and decoder  $G_D\{e_2,e_3,e_4,e_5,e_6,d_1,d_2,d_3,d_4,d_5\}$. The feature map of  $e_1$ layer is defined as latent space $z$, because we use the U-Net structure network and it can not decode latent space if other layers' feature map as latent space $z$. For example, if we define $e_2$ output feature as latent space $z$, then $G_E\{e_1,e_2\}$ and  $G_D\{e_3,e_4,e_5,e_6,d_1,d_2,d_3,d_4,d_5\}$ will as encoder and decoder respectively. We can do linear interpolation $z = \alpha G_E(x_p) + (1 -\alpha){G_E(x_q)}$, but we can not decode latent space $z$, because the  calculation of feature map of $d_5$ requires  the feature map of $e_1$ in U-Net structure, while decoder $G_D$ does not include $e_1$ layer. 
		
		\begin{figure}[htbp]
		\centering
		\includegraphics[height = 2.95cm]{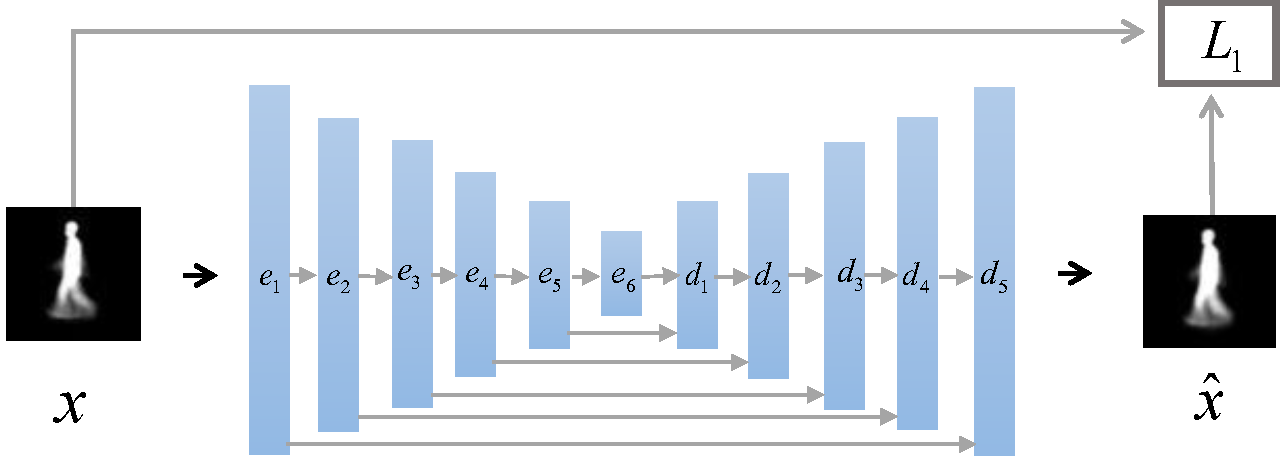}
		\caption{The U-net architecture generator of DV-GAN for modelling gait view space distribution. }
		\label{fig:GAN}
	\end{figure}
	
		\begin{figure}[htbp]
		\centering
		\includegraphics[height = 4.5cm]{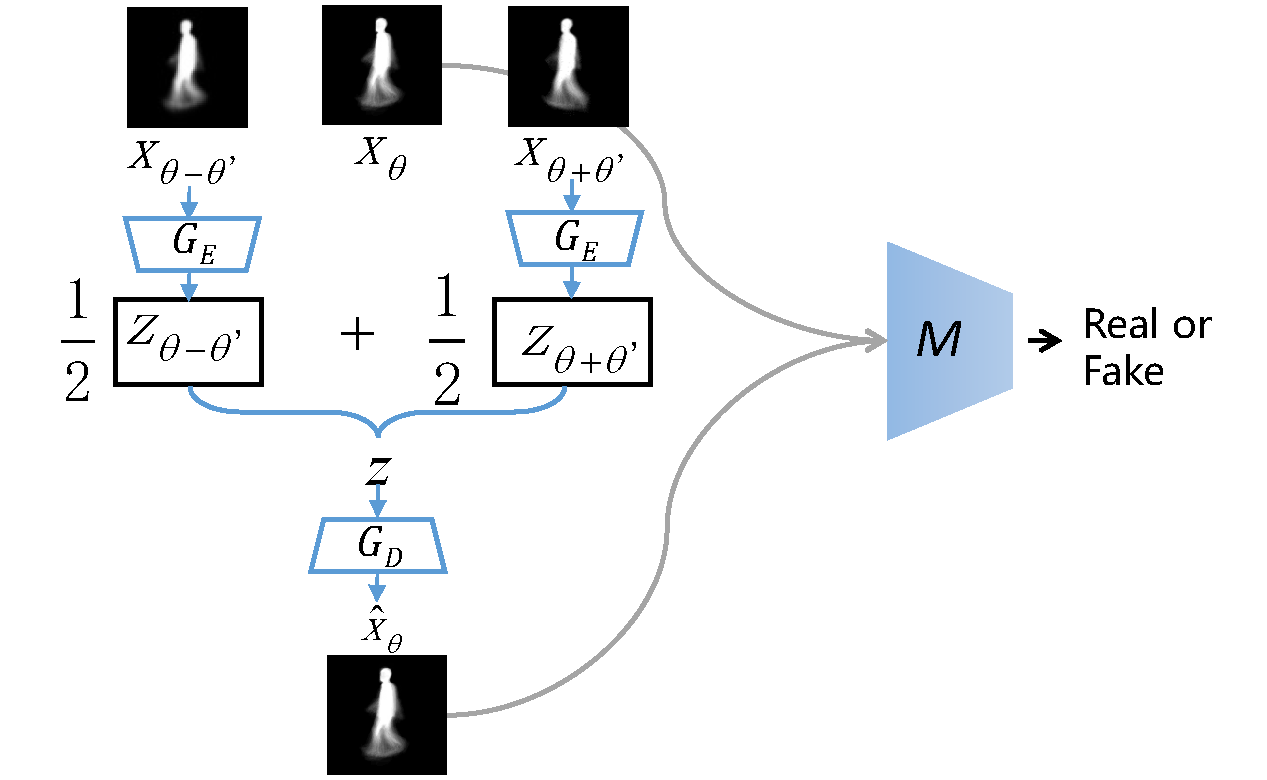}
		\caption{The structure of the real/fake monitor. Monitor will identify if the generate image $\hat{x}_\theta$ is same as the original image $x_\theta$ or not, to preserve human identification and view information.}
		\label{fig:Minitor}
	\end{figure}
	
		\item \textbf{Discriminator}: 	Our discriminator network $D$ ensures that the generated GEIs are realistic. It takes a pair of a real image $x$ and a synthesized image $\hat{x}$ as input, and is trained to identify whether an image is real or not. If the input image is from a real image, the discriminator output value is 1, otherwise 0. The discriminator would  ensure the generated GEI is more and more similar to the original image. The objective of discriminator is:	
		\begin{multline}\label{GAN funtion}
		\mathop{\min}_{G}\mathop{\max}_{D}\mathbb{E}_{x,z~p_{data (x,z)}}[logD (x,x)] +  
		\\
		\mathbb{E}_{x,z~p_{data (x,z)}}[1 - logD (x,G (z))]
		\end{multline}
		
		\item \textbf{Monitor}: 
		Monitor is designed to preserve human identification and view information, as shown in Figure~\ref{fig:Minitor}.
		This idea is inspired by identification-discriminator of GaitGAN~\cite{Yu2017GaitGAN}, which takes a source image and a target image as input and is trained to identify whether the input pair is the same person. Our monitor not only can maintain the identification information, but also can preserve the view information. The monitor has three input images  $x_{\theta-\theta'}, x_\theta$ and $ x_{\theta+\theta'}$. In the training process, first, monitor would generate an image $\hat{x}_\theta$ by decode the latent feature $z$, where latent feature $z$ is the mean value of encoded features ($z_{\theta-\theta'}$ and $z_{\theta+\theta'}$) of two input images ($x_{\theta-\theta'}$ and $ x_{\theta+\theta'}$) . And then the monitor would produce a scalar probability to indicate if the generate image $\hat{x}_\theta$ is same as the original image $x_\theta$ or not, so the identification and view information would be maintained in the training phase, the formula as follows: 	
	
		\begin{multline}\label{preservation}
		\mathop{\min}_{G}\mathop{\max}_{D}\mathbb{E}_{x_\theta,z~p_{data (x_\theta,z)}}[logD (x_\theta,x_\theta)] +  
		\\
		\mathbb{E}_{x_\theta,z~p_{data (x_\theta,z)}}[1 - logD (x_\theta, \hat{x}_\theta)]
		\\
		where \quad \hat{x}_\theta = G(\frac{1}{2}E(x_{\theta-\theta'}) + \frac{1}{2}E(x_{\theta+\theta'}))
		\end{multline}

	\end{itemize}

	\section{Experiments and Analysis}
\subsection{Datasets}
The proposed method is evaluated on  CASIA-B dataset~\cite{yu2006framework} with 11 views and  OU-ISIR Large Population Dataset~\cite{Iwama_IFS2012} with 4 views, respectively.

CASIA-B gait dataset~\cite{yu2006framework} is one of the popular public gait databases and it was created by the Institute of Automation, Chinese Academy of Sciences in January 2005. It contains 124 subjects, and each subject has six sequences.  This set is captured at 11 views, from 0$^\circ$ to 180$^\circ$ with 18$^\circ$ interval between two nearest views. The view angles are \{0$^\circ$, 18$^\circ$, $\cdots$, 180$^\circ$\}. The left image in Figure~\ref{fig:framework} illustrates the samples involving 11 views from a normal walking subject. CASIA-B has three different conditions, normal walking (NM), walking with bag (BG) and walking with a coat (CL). Our proposed method focus on normal walking (NM) condition to solve the view variation. 

In order to better evaluate our method, we perform another experiment on  OU-ISIR Large Population Dataset~\cite{Iwama_IFS2012} with only 4 views ($55^\circ$,$65^\circ$,$75^\circ$ and $85^\circ$).  OU-ISIR is a very large dataset which contains 4007 subjects ranging from 1 to 94 years old. It includes two sequences under the normal walking conditions. It enables us to study the upper bound of gait recognition performance in a more statistically reliable manner.

\subsection{Experimental Setting}
For fair comparison with SPAE~\cite{Yu2017View} and GaitGAN~\cite{Yu2017GaitGAN} later, our experimental setting is same as that of them. The training set contains the first 62 subjects under 6 normal sequences, and the test set contains the rest of subjects. In the test set, the gallery set contains the first 4 normal sequences and the probe set is consists of the rest 2 normal sequences, as shown in Table~\ref{Tab:experiment setting}.  We only synthesize GEIs in training set, not in the test set in our proposed method. Because we want to show that the model trained by dense view synthesized samples can improve performance in the test set compared with original dataset.

\begin{table}[htbp]
	\caption{Experimental setting on CASIA-B dataset (NM: normal walking).}
	\begin{center}
		\begin{tabular}{l|l|l}
			\hline                  
			\multirow{2}{*}{Training}&\multicolumn{2}{c}{Test}\\
			\cline{2-3}
			& Gallery Set & Probe Set \\
			\hline
			ID: 001-062 &ID: 063-124 &ID: 063-124\\
			NM01-NM06 &NM01-NM04&NM05-NM06\\
			\hline                  
			
		\end{tabular}
	\end{center}
	\label{Tab:experiment setting}
\end{table}

The setting of OU-ISIR~\cite{Iwama_IFS2012} is similar to that of CASIA-B. In the experiment, we divide all the subjects into five sets randomly, and keep one set for testing and four sets for training to synthesize samples. In each test set, the first sequence is put into gallery set and the rest sequence is put into probe set.

\subsection{Implementation Details of DV-GAN}
Our structure of DV-GAN is inspired by the idea of Isola {\it et al.}~\cite{pix2pix2017}.  Authors provide an open code, namely {\it pix2pix}, to solve the problem of image-to-image translation. This networks can learn the mapping from input image to output image, which is effective at synthesizing photos. The number of layers of our generator and discriminator is less than that of {\it pix2pix}, because our input size of image is $64\times 64$, while {\it pix2pix} is $256\times 256$. 
The implementation detail of generator and discriminator can be seen in Table~\ref{label.paramer1} and Table~\ref{label.paramer2}. The output of discriminator is one dimensional, the convolution layer 2 is applied to map to a one dimensional output, followed by a sigmoid function.

\begin{table}
	\begin{threeparttable}
		\centering
		\caption{Implementation details of the Generator network}
		\label{label.paramer1}
		\begin{tabular}{|p{1.5cm}<{\centering}@{}p{1.3cm}<{\centering}@{}p{1.7cm}<{\centering}@{}p{1cm}<{\centering}@{}p{1cm}<{\centering}@{}p{1.5cm}<{\centering}|}
			\hline
			\begin{tabular}{c} Layers \end{tabular}&
			\begin{tabular}{c} Number \\of filters \end{tabular}&
			\begin{tabular}{c} Filter size \end{tabular}&
			\begin{tabular}{c} Stride \end{tabular}&
			\begin{tabular}{c} Batch\\ norm \end{tabular}&
			\begin{tabular}{c} Activation\\ function \end{tabular}\\
			\hline
			\hline
			Conv.1 & 64 & 5$\times$ 5 & 2 & N & L-ReLU \\
			Conv.2 & 128 & 5$\times$ 5  & 2 & Y & L-ReLU
			\\
			Conv.3 & 256 & 5$\times$ 5 & 2 & Y & L-ReLU
			\\
			Conv.4 & 512 & 5$\times$ 5 & 2 & Y & L-ReLU
			\\
			Conv.5 & 512 & 5$\times$ 5 & 2 & Y & L-ReLU
			\\
			Conv.6 & 512 & 5$\times$ 5 & 2 & Y & L-ReLU
			\\
			Deconv.1 & 512 & 5$\times$ 5 & 2 & Y & ReLU
			\\
			Deconv.2  & 512 & 5$\times$ 5 & 2 & Y & ReLU
			\\
			Deconv.3  & 256 & 5$\times$ 5 & 2 & Y & ReLU
			\\
			Deconv.4  & 128 & 5$\times$ 5 & 2 & Y & ReLU
			\\
			Deconv.5  & 64 & 5$\times$ 5 & 2 & Y & ReLU
			\\
			Deconv.6  & 64 & 5$\times$ 5 & 2 & N & Tanh
			\\
			\hline
		\end{tabular}
	\end{threeparttable}
	
\end{table}

\begin{table}
	\begin{threeparttable}
		\centering
		\caption{Implementation details of the Discriminator network.}
		\label{label.paramer2}
		\begin{tabular}{|p{1.5cm}<{\centering}@{}p{1.3cm}<{\centering}@{}p{1.7cm}<{\centering}@{}p{1cm}<{\centering}@{}p{1cm}<{\centering}@{}p{1.5cm}<{\centering}|}
			\hline
			\begin{tabular}{c} Layers \end{tabular}&
			\begin{tabular}{c} Number \\of filters \end{tabular}&
			\begin{tabular}{c} Filter size \end{tabular}&
			\begin{tabular}{c} Stride \end{tabular}&
			\begin{tabular}{c} Batch\\ norm \end{tabular}&
			\begin{tabular}{c} Activation\\ function \end{tabular}\\
			\hline
			\hline
			Conv.1 & 64 & 5$\times$ 5  & 2 & N & L-ReLU \\
			Conv.2 & 128 & 5$\times$ 5& 1 & Y & L-ReLU
			\\
			
			\hline
		\end{tabular}
	\end{threeparttable}
\end{table}

\begin{table}
	\begin{threeparttable}
		\centering
		\caption{Implementation details of the Monitor network.}
		\label{label.Monitor}
		\begin{tabular}{|p{1.5cm}<{\centering}@{}p{1.3cm}<{\centering}@{}p{1.7cm}<{\centering}@{}p{1cm}<{\centering}@{}p{1cm}<{\centering}@{}p{1.5cm}<{\centering}|}
			\hline
			\begin{tabular}{c} Layers \end{tabular}&
			\begin{tabular}{c} Number \\of filters \end{tabular}&
			\begin{tabular}{c} Filter size \end{tabular}&
			\begin{tabular}{c} Stride \end{tabular}&
			\begin{tabular}{c} Batch\\ norm \end{tabular}&
			\begin{tabular}{c} Activation\\ function \end{tabular}\\
			\hline
			\hline
			Conv.1 & 64 & 5$\times$ 5  & 2 & N & L-ReLU \\
			Conv.2 & 128 & 5$\times$ 5& 1 & Y & L-ReLU
			\\
			
			\hline
		\end{tabular}
	\end{threeparttable}
\end{table}

In addition, we add additional monitor to preserve  human identification and view information. The implementation detail of monitor (Table~\ref{label.Monitor}) is the same as that of discriminator, but their input data setting is different. The number of input image in discriminator is two, while monitor is three. In the experiment on CASIA-B, we set the $\theta' = 18 ^\circ$ in the Equation~\ref{preservation}, where $\theta$ $\in$ \{18$^\circ$,36$^\circ$,54$^\circ$,72$^\circ$,90$^\circ$,108$^\circ$,126$^\circ$,144$^\circ$,172$^\circ$\}.

After we train the DV-GAN model, DV-GEIs set will be generated. We synthesize GEI from 0$^\circ$ to 180$^\circ$ with 1$^\circ$ interval by linear transformation  $z = \alpha z_p + (1 - \alpha){z_q}$ and decoder latent space $G_D(z)$. Follow the Equation~\ref{synthesize}, we set $\alpha \in \{\frac{1}{18}, \frac{2}{18}, \cdots, \frac{17}{18} \}$, where the angle set of $z_p$ and $z_q$ is $\{(0^\circ,18^\circ),  (18^\circ,36^\circ), \cdots, (162^\circ,180^\circ)\}$. So we can get the various view angle GEIs that do not exist in the original dataset. Finally, we combine the synthesized GEIs and original GEIs to form the DV-GEIs set, and fed into CNN to extract invariant  feature.

\subsection{Implementation Details of Feature Extraction}
We use a simple CNN to extract view invariant feature from DV-GEIs set. We borrow the idea of CNN structure and multi-loss function of PoseGait~\cite{liao2020model} which can effectively extract gait dynamic and static information from human pose sequence. The network details can be seen in Table~\ref{label.paramer3}.

\begin{figure*}[htbp]
	\centering
	\subfigure[]{\includegraphics[width=0.24\textwidth]{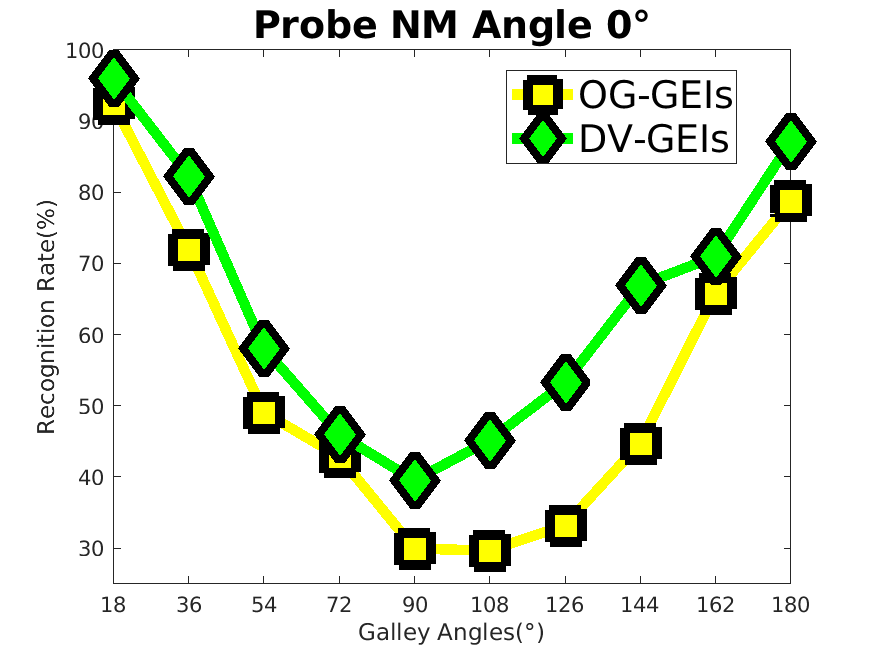}}
	\subfigure[]{\includegraphics[width=0.24\textwidth]{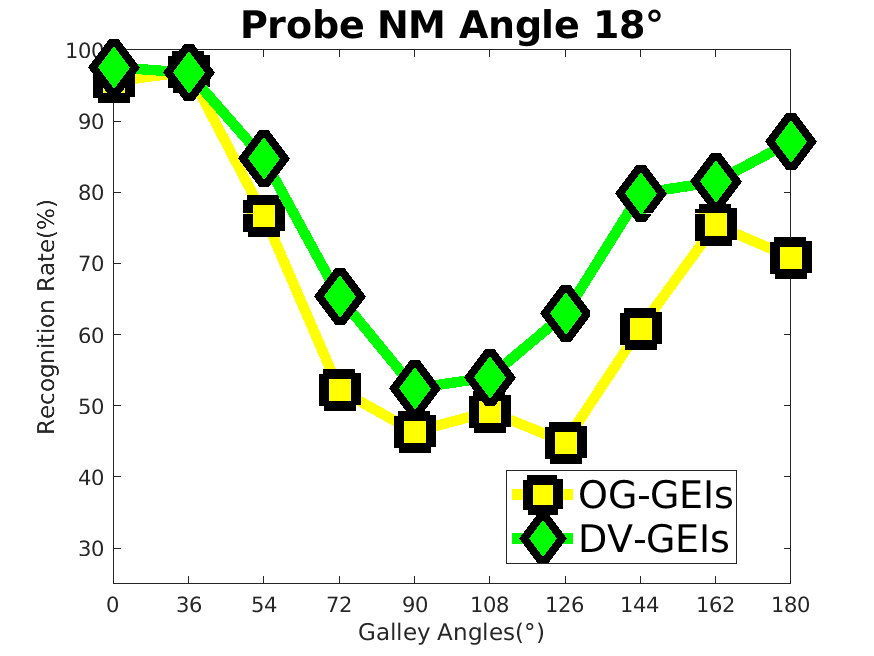}}
	\subfigure[]{\includegraphics[width=0.24\textwidth]{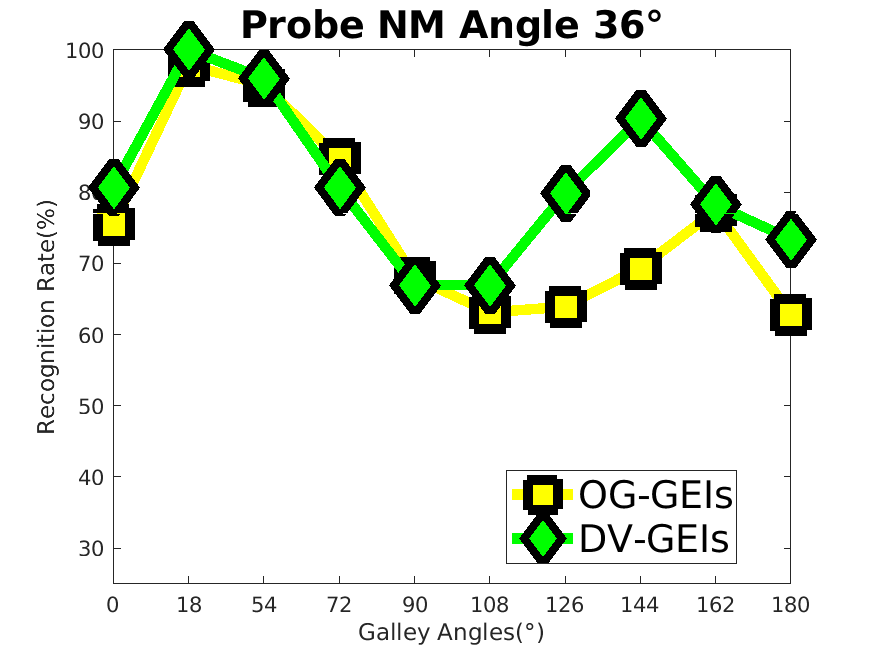}}
	\subfigure[]{\includegraphics[width=0.24\textwidth]{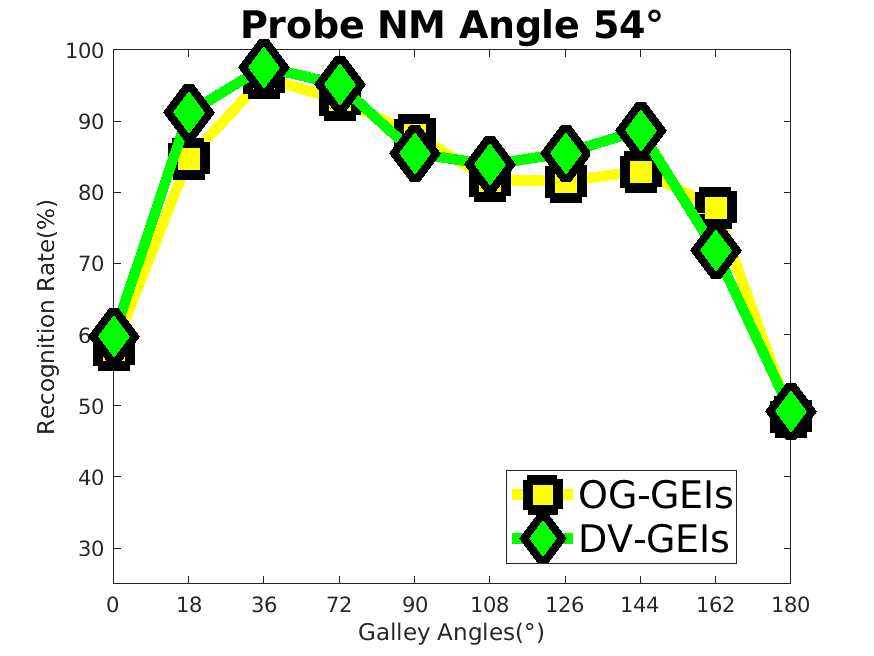}}	
	\subfigure[]{\includegraphics[width=0.24\textwidth]{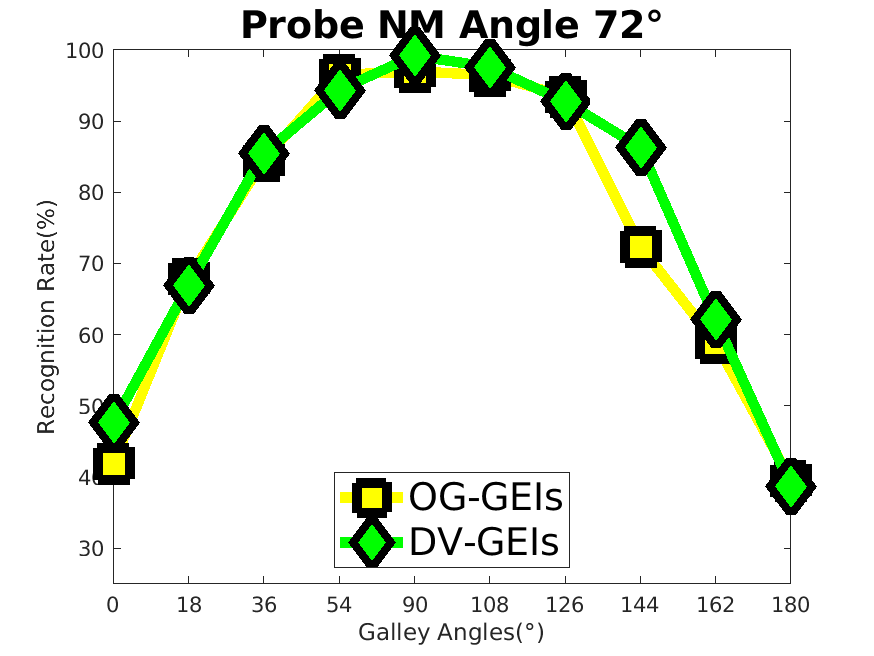}}
	\subfigure[]{\includegraphics[width=0.24\textwidth]{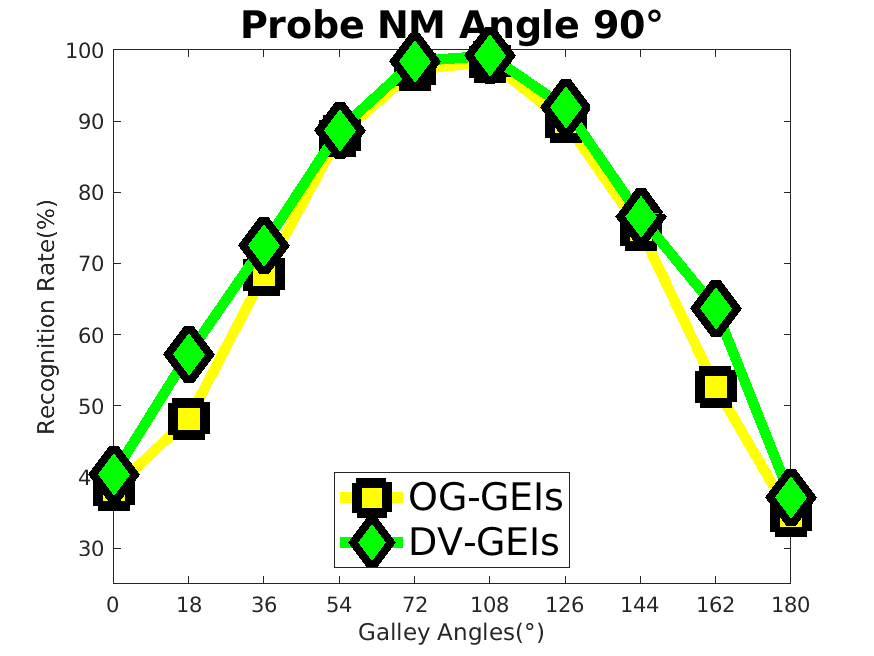}}
	\subfigure[]{\includegraphics[width=0.24\textwidth]{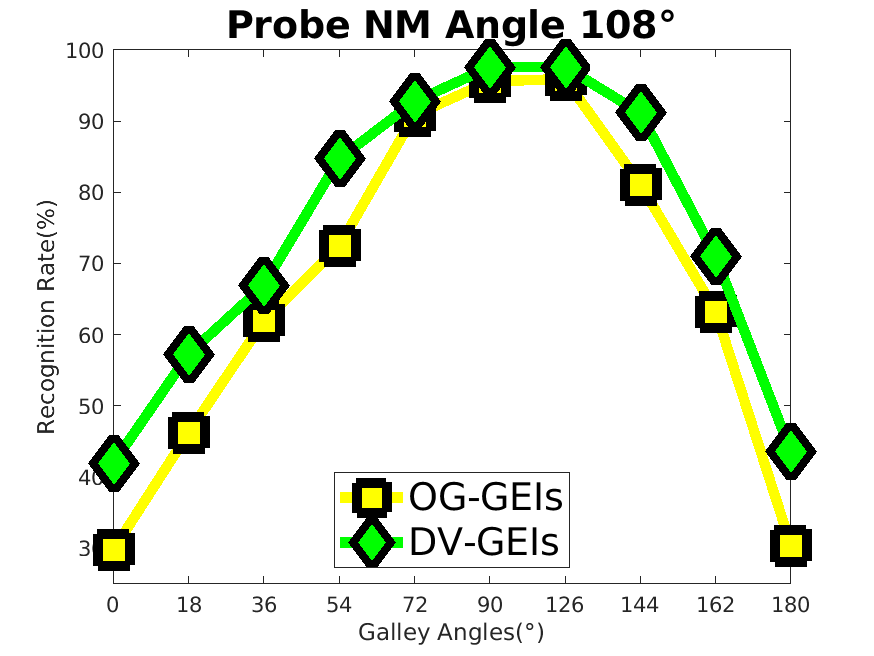}}
	\subfigure[]{\includegraphics[width=0.24\textwidth]{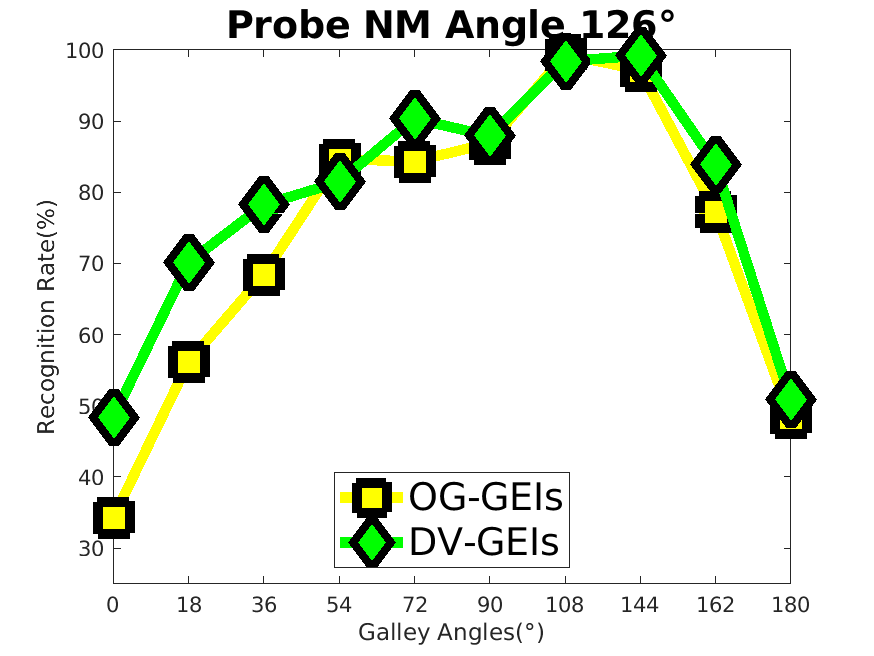}}		
	\subfigure[]{\includegraphics[width=0.24\textwidth]{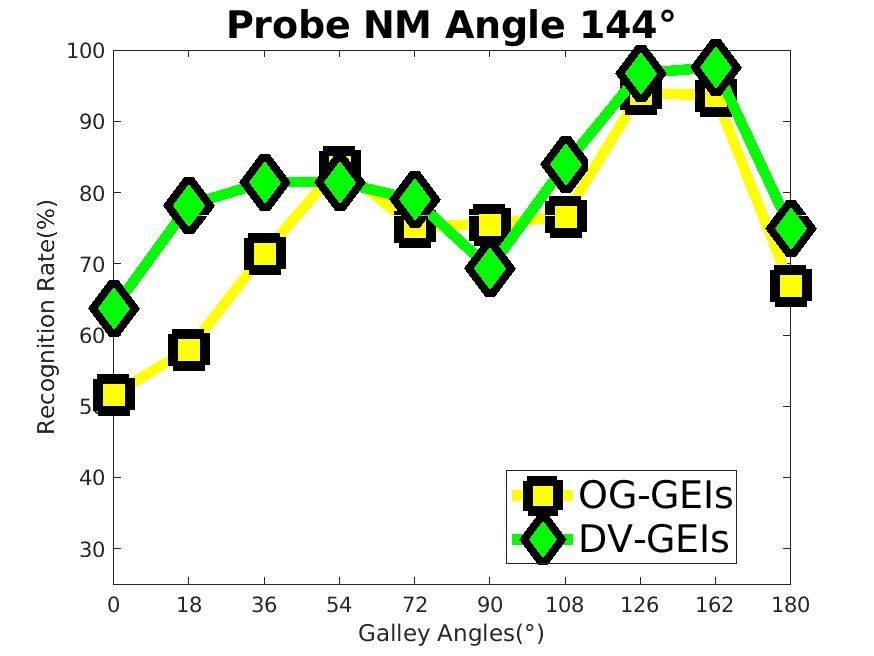}}
	\subfigure[]{\includegraphics[width=0.24\textwidth]{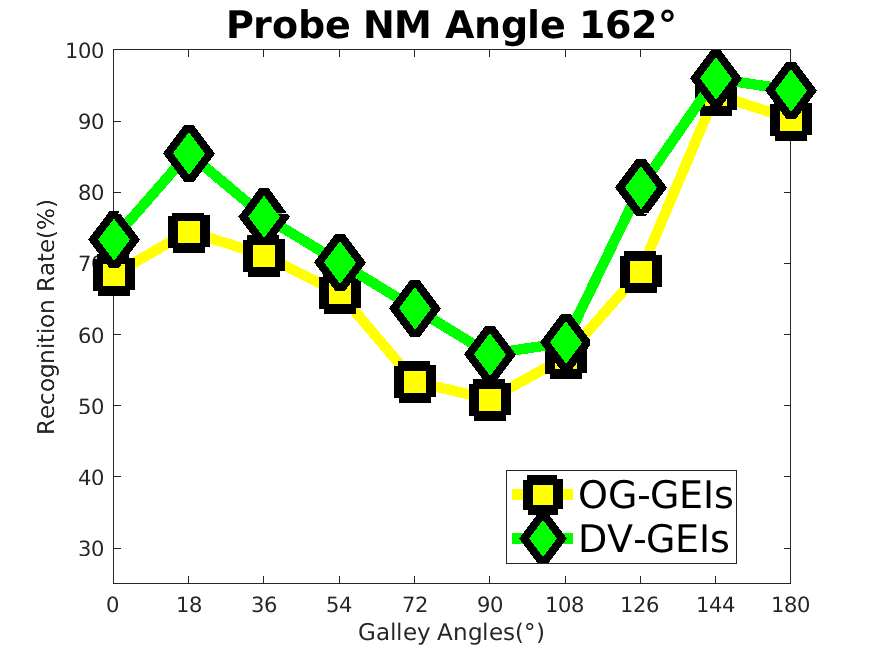}}
	\subfigure[]{\includegraphics[width=0.24\textwidth]{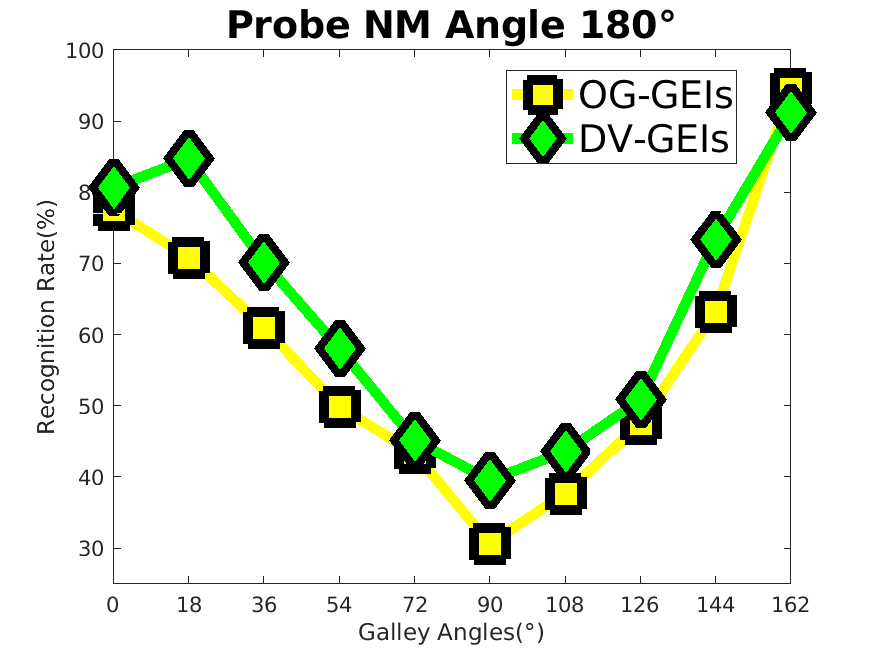}}
	\caption{Comparison with OG-GEIs model that is trained by using original GEIs set, while DV-GEIs model is trained by using proposed dense view set. Each row represents a probe angle and each column represents different probe sequences in the test set. The comparison shows that samples with dense view synthesized by DV-GAN can further learn better view invariant feature. }
	\label{fig.comparing}
\end{figure*}

\begin{figure}[htbp]
	\centering
	\includegraphics[height = 5.5cm]{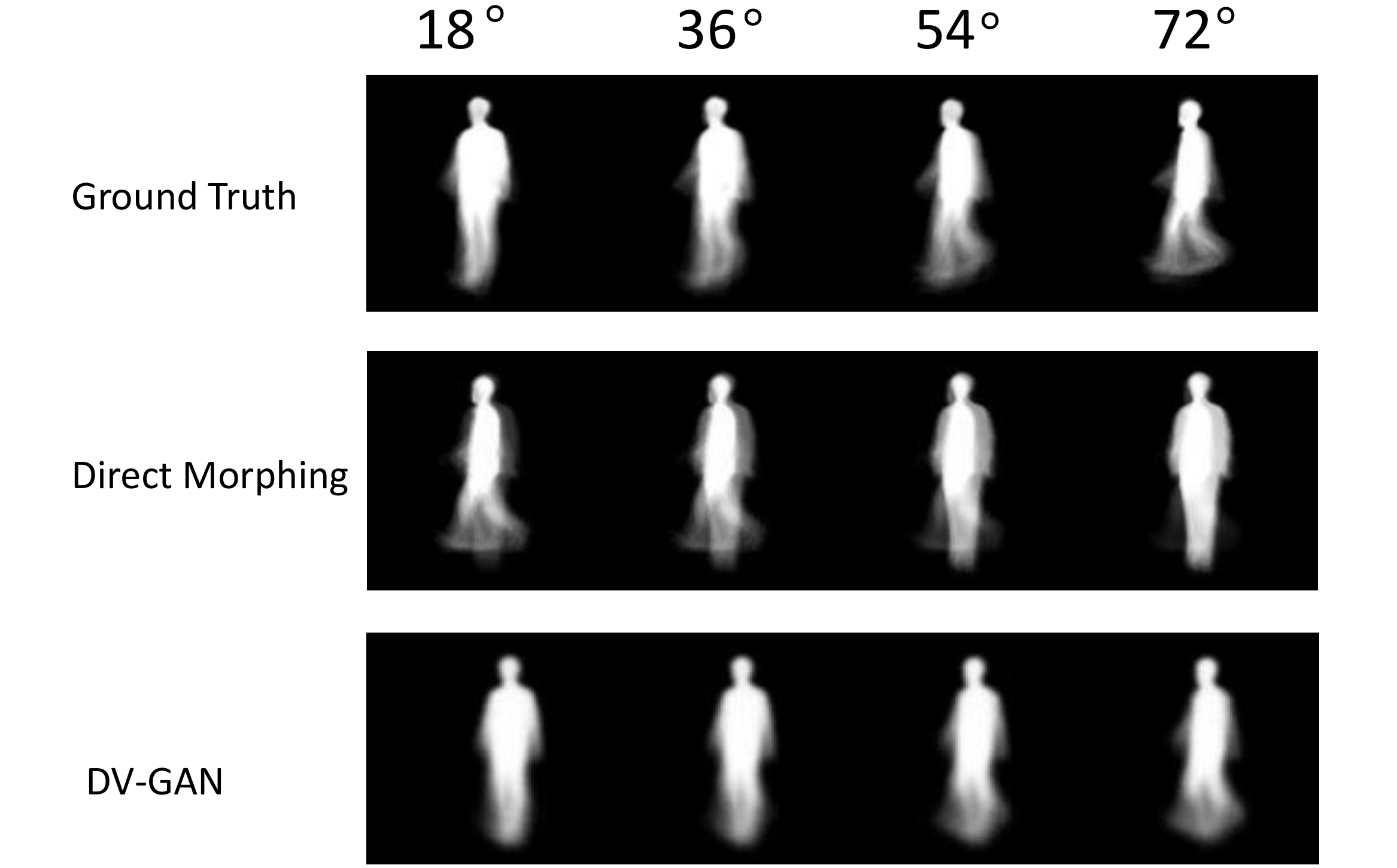}
	\caption{Visualization of synthesized GEIs. Second row GEIs are synthesized by linear interpolation of two GEI images. Third row GEIs are synthesized by proposed DV-GAN. Two types of synthesized GEIs are generated  by 0$^\circ$ and 90$^\circ$ two original GEIs (see the visual demo in supplementary material).}
	\label{fig:compare}
\end{figure}

\begin{table}
	\begin{threeparttable}
		\centering
		\caption{Details implementation of the CNN.}
		\label{label.paramer3}
		\begin{tabular}{|p{1.5cm}<{\centering}@{}p{2cm}<{\centering}@{}p{1.7cm}<{\centering}@{}p{1.4cm}<{\centering}@{}p{1.5cm}<{\centering}|}
			\hline
			\begin{tabular}{c} Layers \end{tabular}&
			\begin{tabular}{c} Number \\of filters \end{tabular}&
			\begin{tabular}{c} Filter size \end{tabular}&
			\begin{tabular}{c} Stride \end{tabular}&
			
			\begin{tabular}{c} Activation\\ function \end{tabular}\\
			\hline
			\hline
			Conv.1 & 32 & 3$\times$ 3  & 1 &  P-ReLU \\
			Conv.2 & 64 & 3$\times$ 3& 1 &  P-ReLU \\
			Pooling.1 & N & 2$\times$ 2& 2 &  N \\
			Conv.3 & 64 & 3$\times$ 3& 1 &  P-ReLU \\
			Conv.4 & 64 & 3$\times$ 3& 1 &  P-ReLU \\
			\hline
			Eltwise.1 &\multicolumn{4}{c|}{Sum operation between Pooling.1 and Conv.4}\\
			\hline
			Conv.5 & 128 & 3$\times$ 3& 1 &  P-ReLU \\
			Pooling.2 & N & 2$\times$ 2& 2 &  N \\
			Conv.6 & 128 & 3$\times$ 3& 1 &  P-ReLU \\
			Conv.7 & 128 & 3$\times$ 3& 1 &  P-ReLU \\
			\hline
			Eltwise.2 &\multicolumn{4}{c|}{Sum operation between Pooling.2 and Conv.7}\\
			\hline
			Conv.8 & 128 & 3$\times$ 3& 1 &  P-ReLU \\
			Conv.9 & 128 & 3$\times$ 3& 1 &  P-ReLU \\
			\hline
			Eltwise.3 &\multicolumn{4}{c|}{Sum operation between Eltwise.2 and Conv.9}\\
			\hline
			Conv.10 & 128 & 3$\times$ 3& 1 &  P-ReLU \\
			FC.1 & 512 & N & N & N \\
			
			\hline
		\end{tabular}
	\end{threeparttable}
\end{table}

Our multi-loss function consists of center loss and softmax loss.  The center loss~\cite{Wen2016A} with the softmax loss jointly supervise the learning of our CNN. The multi-loss function  is defined as Equation~\ref{multi_loss}. 	The softmax loss is useful to pull apart different GEIs and it can enlarge the inter-class dispersion. The center loss would minimizing the intra-class variation and keep the features of different classes separable. In the CNN training process, each batch should calculate several centers by averaging the GEIs features of the corresponding label. 

	\begin{multline}\label{multi_loss}
	L = L_S + \gamma L_c
	\\
	=- \sum_{i=1}^m  \log \frac{e^{W^T_{l_i}\hat{x}_i+b_{l_i}}}{\sum_{j=1}^n e^{W^T_j\hat{x}_i+b_j}}+\frac{\gamma}{2}\sum_{i=1}^m||\hat{x}_i-c_{li}||^2_2
	\end{multline}

	where $\hat{x}_i\in\mathbb{R}^{d} $ is the $i$th GEI feature that belongs to the $l_i$th class. $d$, $W\in\mathbb{R}^{d\times n}$ and $b\in\mathbb{R}^{d}$ denote the feature dimension, last connected layer and bias term, respectively.
	$c_{li}\in\mathbb{R}^{d}  $ is the $l_i$th class center of gait features. 	We set $\gamma=0.008$ in the experiment.

\subsection{Experimental Results on CASIA-B dataset}
The experimental results on CASIA-B dataset are shown in Table~\ref{label.probenm_GAN}. In this table, each row is correspondent to a GEI angle of the gallery, and each column is correspondent to the angle of the probe set. The recognition rate of the cross-view condition has 121 combinations.

In order to illustrate synthesized samples by our DV-GAN can make contributions to the improvement of gait recognition, we compare with another experiment OG-GEIs result.  OG-GEIs model is trained by using original GEIs set. From Figure~\ref{fig.comparing}, we can see performance of DV-GEIs is better than OG-GEIs at many points. This shows that our dense view samples synthesized by DV-GAN is an effective way to learn better view invariant feature and enhance the robustness  for gait recognition.

\subsection{Visualization of Synthesized GEIs} \label{Visualization}
In order to see the quality of generated image, we synthesize some view samples, as shown in Figure~\ref{fig:compare}. In fact, in our above experiment, we generate the GEIs by using two adjacent GEIs with 18$^\circ$ interval in CASIA B dataset. However, the difference between the adjacent angle (18$^\circ$) is hard to be distinguished in vision, and without ground truth GEIs in the original dataset. In order to visualize the transformation obviously and have ground truth for comparison, we use two GEIs (0$^\circ$ and 90$^\circ$) with large 90$^\circ$ interval to generate some sample GEIs. That is, in the linear transformation $z = \alpha z_p + (1 - \alpha){z_q}$, we set linear ratio $\alpha \in \{ \frac{18}{90}, \frac{36}{90}, \frac{54}{90}, \frac{72}{90} \}$, and the angle set of $z_p$ and $z_q$  is $\{(0^\circ,90^\circ)\}$.

To better show the contributions of DV-GAN, we compare our DV-GEIs with another type synthesized GEIs which direct view morphing by linear interpolation of two GEI images. That is, the second row GEIs are generated by equation $\hat{x} = \alpha x_p + (1 - \alpha){x_q}$, where $\alpha \in \{ \frac{18}{90}, \frac{36}{90}, \frac{54}{90}, \frac{72}{90} \}$. From Figure~\ref{fig:compare}, we can see  that synthesized GEIs by direct view morphing have obvious ghost, while synthesized GEIs by DV-GANs is very similar to ground truth although they are synthesized by two images with $90^\circ$ interval. It will be more similar to the original image if using images with a smaller angle interval to synthesize.  The comparison shows that our DV-GAN can synthesize realistic image with any angle condition, and our generated image can preserve human view information very well.

\begin{table*}[htbp]
	\centering
	\caption{Recognition rates of proposed method.(NM: normal walking).}
	\label{label.probenm_GAN}
   \begin{tabular}{|p{0.5cm}|p{0.9cm}|p{0.9cm}|p{0.9cm}|p{0.9cm}|p{0.9cm}|p{0.9cm}|p{0.9cm}|p{0.9cm}|p{0.9cm}|p{0.9cm}|p{0.9cm}|p{0.9cm}|}
		\hline & & \multicolumn{11}{c|}{Probe set view (NM05,NM06)}\\
		\hline & &0$^\circ$&18$^\circ$&36$^\circ$&54$^\circ$&72$^\circ$&90$^\circ$&108$^\circ$&126$^\circ$&144$^\circ$&162$^\circ$&180$^\circ$\\
		\cline{1-13}\multirow{11}{*}{\rotatebox{90}{Gallery set view (NM01-NM04)}}
		&0$^\circ$&100.0&97.58&80.65&59.68&47.58&40.32&41.94&48.39&63.71&73.39&80.65\\
		\cline{2-13}&18$^\circ$&95.97&100.0&100.0&91.13&66.94&57.26&57.26&70.16&78.23&85.48&84.68\\
		\cline{2-13}&36$^\circ$&82.26&96.77&98.39&97.58&85.48&72.58&66.94&78.23&81.45&76.61&70.16\\
		\cline{2-13}&54$^\circ$&58.06&84.68&95.97&97.58&94.35&88.71&84.68&81.45&81.45&70.16&58.06\\
		\cline{2-13}&72$^\circ$&45.97&65.32&80.65&95.16&99.19&98.39&92.74&90.32&79.03&63.71&45.16\\
		\cline{2-13}&90$^\circ$&39.52&52.42&66.94&85.48&99.19&99.19&97.58&87.90&69.35&57.26&39.52\\
		\cline{2-13}&108$^\circ$&45.16&54.03&66.94&83.87&97.58&99.19&99.19&98.39&83.87&58.87&43.55\\
		\cline{2-13}&126$^\circ$&53.23&62.90&79.84&85.48&92.74&91.94&97.58&97.58&96.77&80.65&50.81\\
		\cline{2-13}&144$^\circ$&66.94&79.84&90.32&88.71&86.29&76.61&91.13&99.19&99.19&95.97&73.39\\
		\cline{2-13}&162$^\circ$&70.97&81.45&78.23&71.77&62.10&63.71&70.97&83.87&97.58&98.39&91.13\\
		\cline{2-13}&180$^\circ$&87.10&87.10&73.39&49.19&38.71&37.10&43.55&50.81&75.00&94.35&97.58\\
		\hline
	\end{tabular}
\end{table*}
\begin{table*}[htbp]
	\centering
	\caption{Comparison with based on GEI template methods on CASIA-B dataset at average accuracy(\%). Excluding identical-view cases.}
	\label{label.compareNewmethod}
	\begin{tabular}{|p{1.2cm}|p{2.3cm}|p{0.5cm}|p{0.5cm}|p{0.5cm}|p{0.5cm}|p{0.5cm}|p{0.5cm}|p{0.5cm}|p{0.5cm}|p{0.5cm}|p{0.5cm}|p{0.5cm}|p{0.7cm}|}
		\hline
		\multirow{2}{*}{\shortstack{Training\\ Subjects}} & \multirow{2}{*}{Methods} & \multicolumn{12}{c|}{Probe angle} \\
		\cline{3-14}
	    & &0$^\circ$&18$^\circ$&36$^\circ$&54$^\circ$&72$^\circ$&90$^\circ$&108$^\circ$&126$^\circ$&144$^\circ$&162$^\circ$&180$^\circ$  & Mean\\ \hline
	    
		\multirow{4}{*}{62}&SPAE~\cite{Yu2017View} &50.0& 58.1& 61.0& 63.3& 64.0& 62.1& 62.3& 66.3& 64.4& 54.5& 46.7& 59.3 \\ 
		\cline{2-14}&GaitGAN~\cite{Yu2017GaitGAN} & 41.9& 53.5& 63.0& 64.5& 63.1& 58.1& 61.7& 65.7& 62.7& 54.1& 40.6& 57.2 	\\ 
		\cline{2-14}&GaitGANv2~\cite{yu2019gaitganv2} & 48.1& 61.9& 68.7& 71.7& 66.7& 64.8& 66.0& 70.2& 71.6& 58.9& 46.1& 63.1\\ 
		\cline{2-14}&DV-GEIs (Ours) & 64.5& 76.2& 81.3& 80.8& 77.1& 72.6& 74.4& 78.9& 80.6& 75.6& 63.7& \textbf{75.1}\\ 
		\hline

		\multirow{2}{*}{74} &GaitSet-GEI~\cite{chao2019gaitset}&-&-&-&-&-&-&-&-&-&-&-& 80.4 \\ 
		\cline{2-14}&DV-GEIs (Ours)  & 71.0 & 86.4& 91.4& 89.6& 80.4& 80.1& 82.5& 90.1& 90.4& 85.3& 70.5& \textbf{83.4}\\ 
		\hline
	\end{tabular}
	
\end{table*}

\subsection{Comparison with VTM methods}\label{Comparison with VTM methods}
\begin{figure}[htbp]
  \centering
  \subfigure[]{\includegraphics[width=0.53\textwidth]{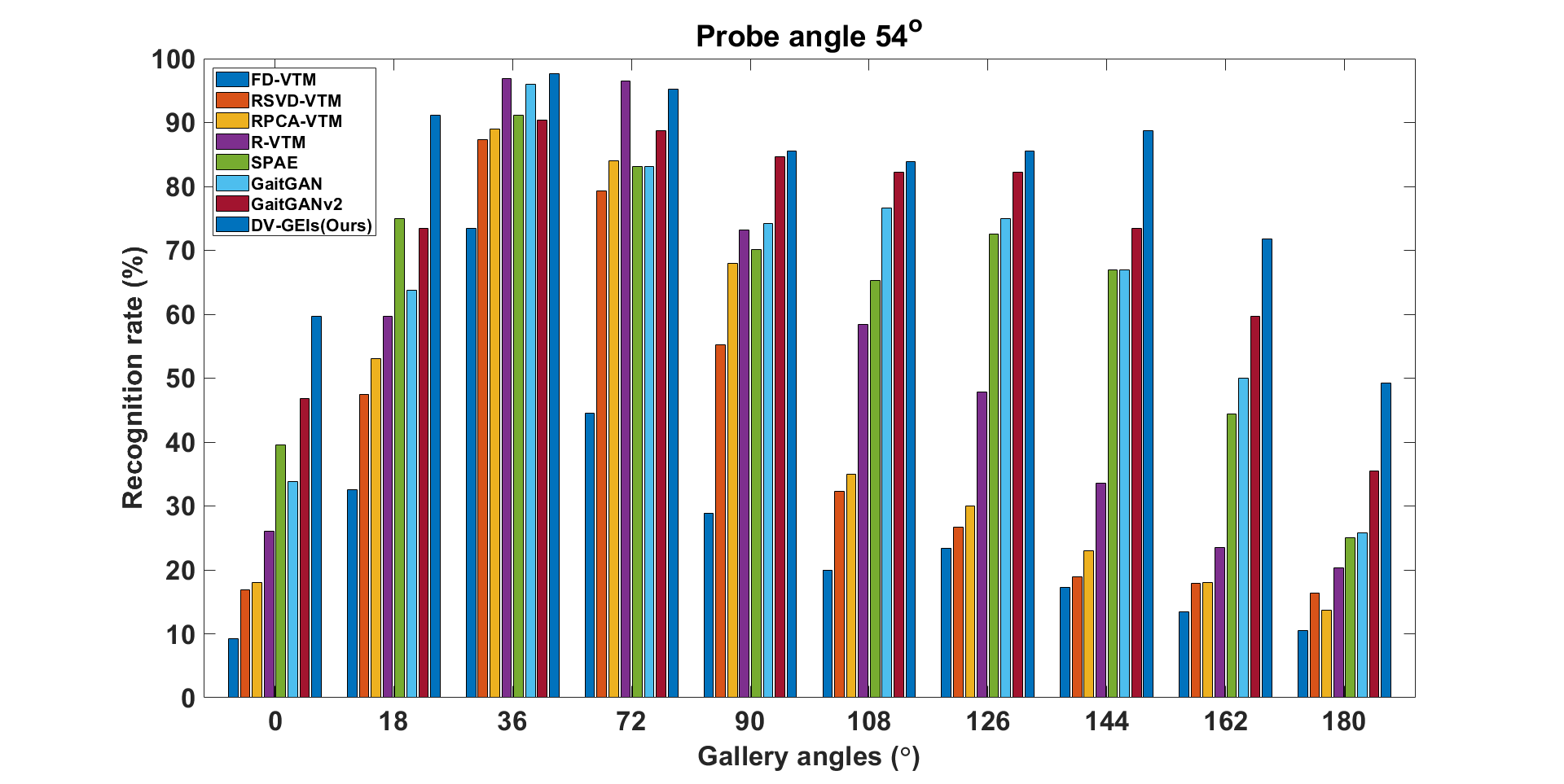}}
  \subfigure[]{\includegraphics[width=0.53\textwidth]{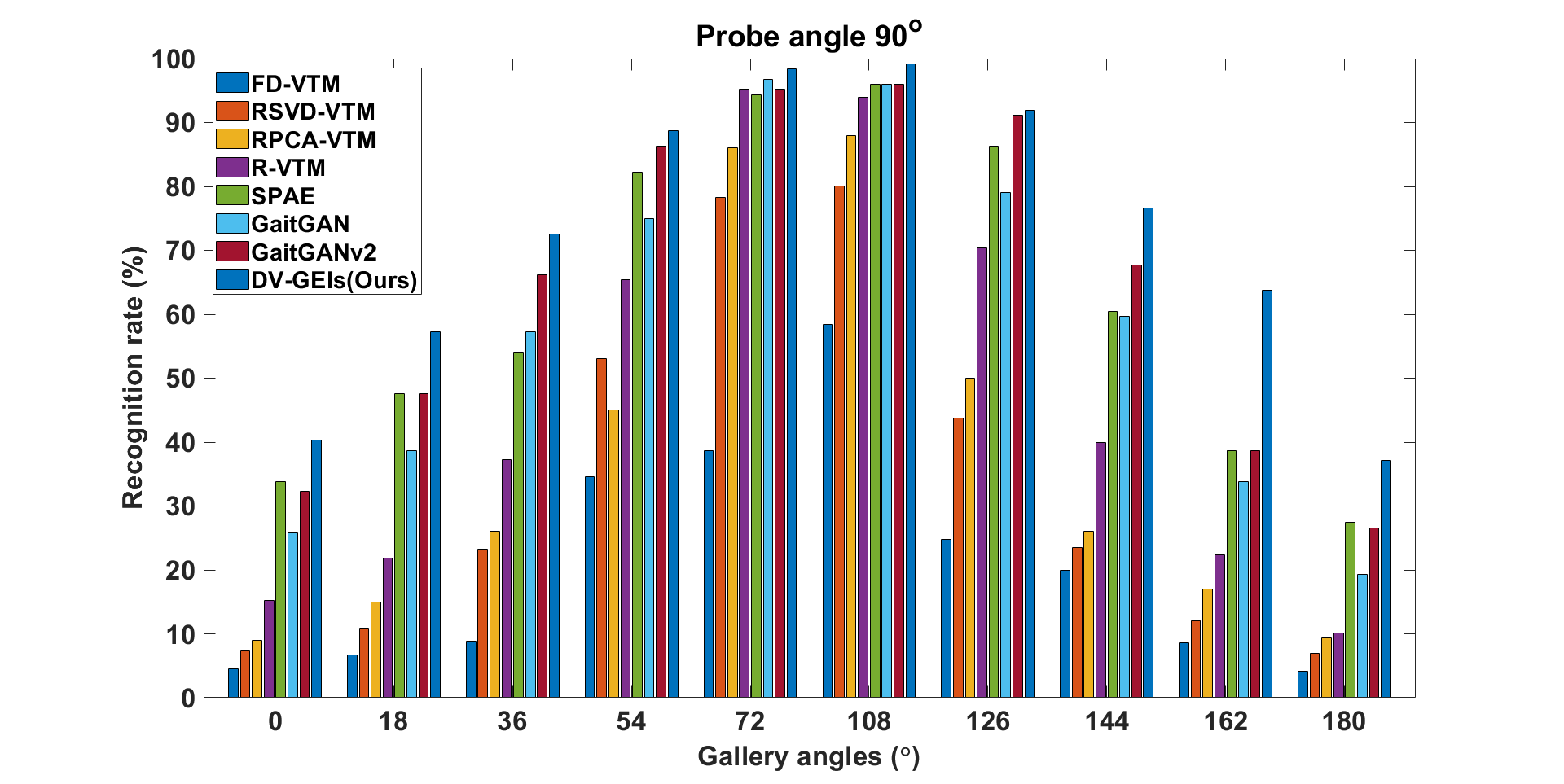}}
  \hfil
  \subfigure[]{\includegraphics[width=0.53\textwidth]{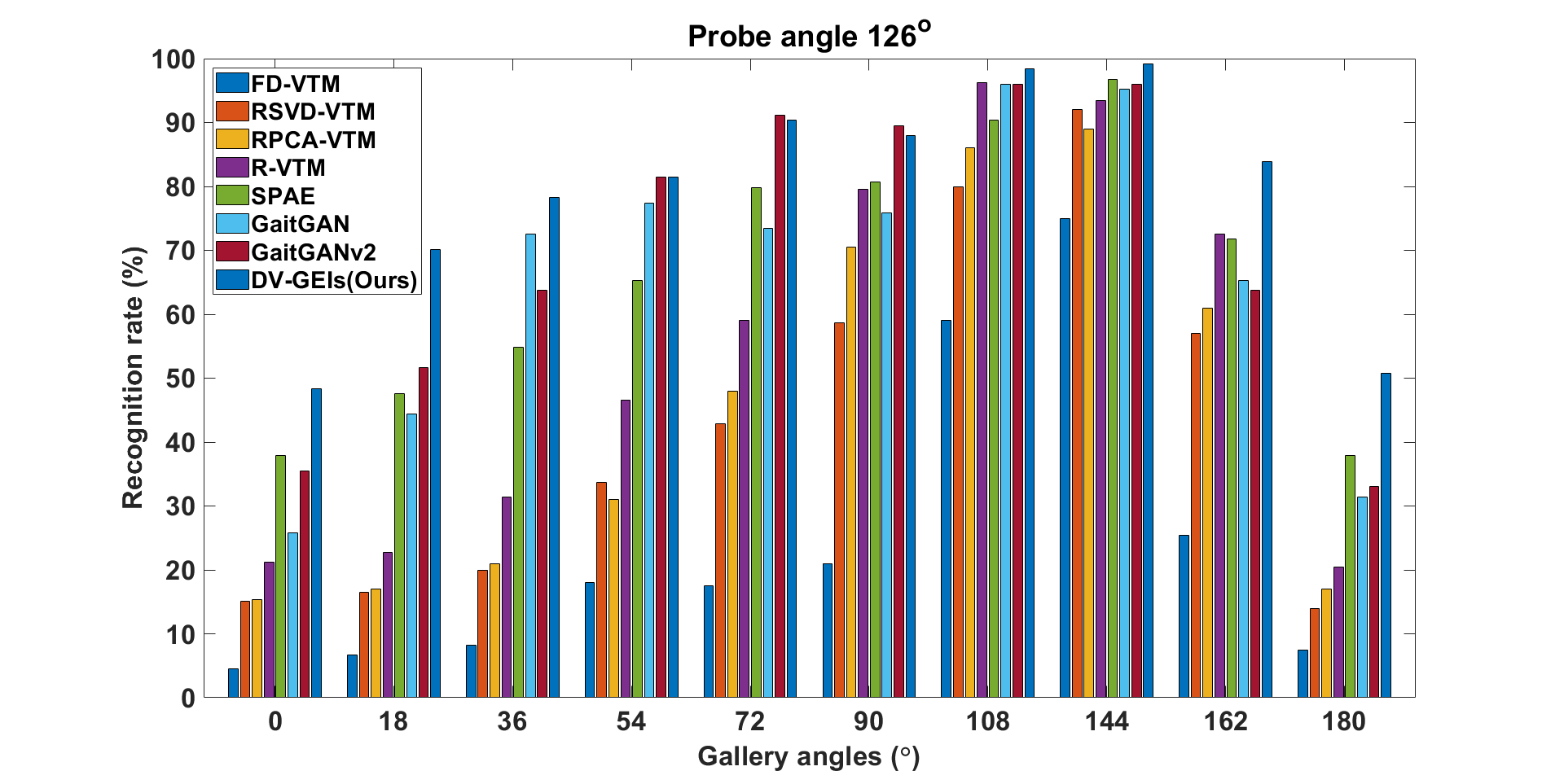}}
  \caption{Comparisons with view transformation model methods at probe angles (a)$54^\circ$, (b)$90^\circ$ and (c)$126^\circ$. The gallery angles are the rest 10 angles except the corresponding probe angle. Our result outperforms those VTM methods. }
  \label{fig.comparingThreeAngles}
\end{figure}

In other to show the advantages of dense view samples, we compare with view transformation model (VTM) methods, including FD-VTM~\cite{makihara:fd-vtm}, RSVD-VTM~\cite{kusa:rsvd-vtm}, RPCA-VTM~\cite{zheng:rpca-vtm}, R-VTM~\cite{kusa:r-vtm}, SPAE~\cite{Yu2017View}, GaitGAN~\cite{Yu2017GaitGAN} and GaitGANv2~\cite{yu2019gaitganv2}, as shown in Figure~\ref{fig.comparingThreeAngles}. Those methods are all trying to transform gait features from one view to another view, while our method is synthesizing more samples to cover the whole view space. 	

The probe angles selected are $54^\circ$, $90^\circ$ and $126^\circ$ in experiments of those methods. From Figure~\ref{fig.comparingThreeAngles}, we can see that the performance of proposed DV-GEIs method outperforms that of others, especially when the angle difference between the gallery and the probe is large. This shows that dense view samples can deal with large viewpoint variation well. In addition, the proposed method can also improve the recognition rate obviously when the viewpoint variation is not large enough.

\subsection{Comparison with GEI template methods}
To further evaluate the proposed method, we compare with based on recent GEI template methods because our input data is also based on GEI, including SPAE~\cite{Yu2017View}, GaitGAN~\cite{Yu2017GaitGAN}, GaitGANv2~\cite{yu2019gaitganv2}, and GaitSet-GEI~\cite{chao2019gaitset}. The comparison as shown in Table~\ref{label.compareNewmethod}.	Compared with those based on GEI template methods, our proposed method mean accuracy (75.1\%) is much better than that of SPAE~\cite{Yu2017View} (59.3\%), GaitGAN~\cite{Yu2017GaitGAN} (57.2\%) and GaitGANv2~\cite{yu2019gaitganv2} (63.1\%). Those methods transform transform any GEI into the side GEI, while ours synthesize dense view samples to cover the whole view space. This show that gait view space covering can better handle with cross-view problem.

The method of GaitSet~\cite{chao2019gaitset} can achieve a very high performance when it uses the human silhouette sequence as input feature. But its performance would be decreased dramatically when use GEI as input feature (drop from 95.0\% to 80.4\%). One reason for this is that human silhouette sequence has much rich information than GEI. Here, we not compare with those based on human silhouette sequence method. Because our method is based on the GEI template, so it maybe unfair to compare with those based on human silhouette sequence method.

\subsection{Experimental results on OU-ISIR dataset}\label{OU-ISIR dataset}
OU-ISIR dataset~\cite{Iwama_IFS2012} is also employed to evaluate the proposed method. In the training phase, the view angle of DV-GEIs on OU-ISIR is from $55^\circ$ to $85^\circ$ with $1^\circ$ interval. This is because start angle if from  $55^\circ$, and end with $85^\circ$ on OU-ISIR dataset.  In the test phase, probe angle and gallery angle are $55^\circ$, $65^\circ$, $75^\circ$ and $85^\circ$, respectively. The result of experiments on OU-ISIR dataset is shown in Table~\ref{label.OU-ISIR}. In this table, each row is correspondent to a angle of gallery set, and each column is correspondent to a angle of probe set. The recognition rate of the cross-view condition has 16 combinations.

We compare our results with OG-GEIs (trained by original GEIs set), DeepCNN~\cite{Wu2017A} and GaitGANv2~\cite{yu2019gaitganv2}, as shown in Table~\ref{label.Comparison-OU-ISIR}. In this table, each column is correspondent to the angle of the probe set. The recognition rate is by averaging different gallery angle, excluding identical view cases. From that table, we can see the performance of DV-GEIs is better than the baseline reported by the dataset authors~\cite{Wu2017A, yu2019gaitganv2} when probe angle is $55^\circ$ and $65^\circ$. In addition, the accuracy DV-GEIs outperforms that of OG-GEIs, which shows again that dense view samples synthesized by DV-GEIs can further improve the robustness to view variation. 

\begin{table}
	\centering
	\caption{Experimental results on OU-ISIR dataset. Model is trained by using DV-GEIs dataset.}
	\label{label.OU-ISIR}
	\begin{tabular}{|c|c|c|c|c|}
		\hline
		\multirow{2}{*}{Probe angle} & \multicolumn{4}{c|}{Gallery angle} \\
		\cline{2-5}
		& $55^\circ$ & $65^\circ$ & $75^\circ$ & $85^\circ$  \\ \hline
		$55^\circ$  & 96.6 &   95.2 &94.5 &88.1 \\ \hline
		$65^\circ$ &  97.3 & 96.8 &94.8 &93.2  \\  \hline
		$75^\circ$  & 90.1 &96.3 & 97.0 &96.8  \\ \hline
		$85^\circ$  & 91.1  &96.3 &96.4 & 96.5  \\ \hline
	\end{tabular}
\end{table}

\begin{table}
	\centering
	\caption{Comparison with other methods on OU-ISIR with average accuracy(\%). Excluding identical view cases. OG-GEIs: trained by using original GEIs set. DV-GEIs: trained by using proposed dense view set. }
	\label{label.Comparison-OU-ISIR}
	\begin{tabular}{|c|c|c|c|c|}
		\hline
		\multirow{2}{*}{\shortstack{Methods}} & \multicolumn{4}{c|}{Probe angle} \\
		\cline{2-5}
		& $55^\circ$ & $65^\circ$ & $75^\circ$ & $85^\circ$  \\ \hline
		DeepCNN~\cite{Wu2017A} & 91.6 & 92.3 & 92.4 & 94.8 \\ \hline
		
		GaitGANv2~\cite{yu2019gaitganv2} & 91.9 & 95.0 & 94.4 & 94.6 \\ \hline
		OG-GEIs(Ours) &  92.0 & 93.8 & 94.0 & 93.8 \\  \hline
		DV-GEIs(Ours)  &{\bf92.6} & {\bf95.1} & 94.4 & 94.6 \\ \hline

	\end{tabular}
\end{table}

\section{Conclusions}
In this paper, we introduce a novel Dense-View GEIs Set (DV-GEIs) to handle with the challenge of samples with limited view angles  on existing gait datasets.   View angle  of DV-GEIs set can cover the whole view space, range from $0^\circ$ to $180^\circ$ with $1^\circ$ interval. This set is synthesized by proposed DV-GAN, which consists of generator, discriminator and monitor. Monitor can preserve human identification and view information very well. The experimental result shows that samples with dense view  can learn better view invariant feature compare with original dataset.

With the development of synthesized sample technology, we believe the idea of dense view samples synthesized by DV-GAN not only can enhance robustness to view variation, but also deal with other variations, like carrying and clothing condition. Eventually, it would  further improve the development of gait recognition. 
 
\section*{Acknowledgment}
This work was  supported in part by NSF I/UCRC grant 1747751.

{\small
\bibliographystyle{ieee}
\bibliography{main}

\begin{thebibliography}{10}\itemsep=-1pt

\bibitem{an2018pose}
W.~An, R.~Liao, S.~Yu, Y.~Huang, and P.~C. Yuen.
\newblock Improving gait recognition with 3d pose estimation.
\newblock In {\em the 13th Chinese Conference on Biometric Recognition}, pages
  137--147, 2018.

\bibitem{an2020performance}
W.~An, S.~Yu, Y.~Makihara, X.~Wu, C.~Xu, Y.~Yu, R.~Liao, and Y.~Yagi.
\newblock Performance evaluation of model-based gait on multi-view very large
  population database with pose sequences.
\newblock {\em IEEE Transactions on Biometrics, Behavior, and Identity
  Science}, 2020.

\bibitem{chao2019gaitset}
H.~Chao, Y.~He, J.~Zhang, and J.~Feng.
\newblock Gaitset: Regarding gait as a set for cross-view gait recognition.
\newblock In {\em Proceedings of the AAAI Conference on Artificial
  Intelligence}, pages 8126--8133, 2019.

\bibitem{chen2018image}
J.~Chen, J.~Chen, H.~Chao, and M.~Yang.
\newblock Image blind denoising with generative adversarial network based noise
  modeling.
\newblock In {\em Proceedings of the IEEE Conference on Computer Vision and
  Pattern Recognition}, pages 3155--3164, 2018.

\bibitem{Goodfellow2014GAN}
I.~J. Goodfellow, J.~Pougetabadie, M.~Mirza, B.~Xu, D.~Wardefarley, S.~Ozair,
  A.~Courville, Y.~Bengio, Z.~Ghahramani, and M.~Welling.
\newblock Generative adversarial nets.
\newblock {\em Advances in Neural Information Processing Systems},
  3:2672--2680, 2014.

\bibitem{Han2006Individual}
J.~Han and B.~Bhanu.
\newblock Individual recognition using gait energy image.
\newblock {\em IEEE Transactions on Pattern Analysis \& Machine Intelligence},
  28(2):316--22, 2006.

\bibitem{hou2017deep}
X.~Hou, L.~Shen, K.~Sun, and G.~Qiu.
\newblock Deep feature consistent variational autoencoder.
\newblock In {\em 2017 IEEE Winter Conference on Applications of Computer
  Vision}, pages 1133--1141, 2017.

\bibitem{pix2pix2017}
P.~Isola, J.-Y. Zhu, T.~Zhou, and A.~A. Efros.
\newblock Image-to-image translation with conditional adversarial networks.
\newblock In {\em Proceedings of the IEEE Conference on Computer Vision and
  Pattern Recognition}, pages 1125--1134, 2017.

\bibitem{Iwama_IFS2012}
H.~Iwama, M.~Okumura, Y.~Makihara, and Y.~Yagi.
\newblock The ou-isir gait database comprising the large population dataset and
  performance evaluation of gait recognition.
\newblock {\em IEEE Transactions on Information Forensics and Security},
  7(5):1511--1521, 2012.

\bibitem{kusa:rsvd-vtm}
W.~Kusakunniran, Q.~Wu, H.~Li, and J.~Zhang.
\newblock Multiple views gait recognition using view transformation model based
  on optimized gait energy image.
\newblock In {\em ICCV Workshops}, pages 1058--1064, 2009.

\bibitem{Kusakunniran2010Multiple}
W.~Kusakunniran, Q.~Wu, H.~Li, and J.~Zhang.
\newblock Multiple views gait recognition using view transformation model based
  on optimized gait energy image.
\newblock In {\em IEEE International Conference on Computer Vision Workshops},
  pages 1058--1064, 2010.

\bibitem{kusa:r-vtm}
W.~Kusakunniran, Q.~Wu, J.~Zhang, and H.~Li.
\newblock Gait recognition under various viewing angles based on correlated
  motion regression.
\newblock {\em IEEE TCSVT}, 22(6):966--980, 2012.

\bibitem{liao2017pose}
R.~Liao, C.~Cao, E.~B. Garcia, S.~Yu, and Y.~Huang.
\newblock Pose-based temporal-spatial network (ptsn) for gait recognition with
  carrying and clothing variations.
\newblock In {\em the 12th Chinese Conference on Biometric Recognition}, pages
  474--483, 2017.

\bibitem{liao2020model}
R.~Liao, S.~Yu, W.~An, and Y.~Huang.
\newblock A model-based gait recognition method with body pose and human prior
  knowledge.
\newblock {\em Pattern Recognition}, 98:107069, 2020.

\bibitem{Makihara2006Gait}
Y.~Makihara, R.~Sagawa, Y.~Mukaigawa, T.~Echigo, and Y.~Yagi.
\newblock Gait recognition using a view transformation model in the frequency
  domain.
\newblock In {\em Proceedings of the European Conference on Computer Vision},
  pages 151--163, 2006.

\bibitem{makihara:fd-vtm}
Y.~Makihara, R.~Sagawa, Y.~Mukaigawa, T.~Echigo, and Y.~Yagi.
\newblock Gait recognition using a view transformation model in the frequency
  domain.
\newblock In {\em ECCV}, pages 151--163, 2006.

\bibitem{qian2018pose}
X.~Qian, Y.~Fu, T.~Xiang, W.~Wang, J.~Qiu, Y.~Wu, Y.-G. Jiang, and X.~Xue.
\newblock Pose-normalized image generation for person re-identification.
\newblock In {\em Proceedings of the European Conference on Computer Vision},
  pages 650--667, 2018.

\bibitem{ronneberger2015u}
O.~Ronneberger, P.~Fischer, and T.~Brox.
\newblock U-net: Convolutional networks for biomedical image segmentation.
\newblock In {\em International Conference on Medical image computing and
  computer-assisted intervention}, pages 234--241. Springer, 2015.

\bibitem{takemura2018multi}
N.~Takemura, Y.~Makihara, D.~Muramatsu, T.~Echigo, and Y.~Yagi.
\newblock Multi-view large population gait dataset and its performance
  evaluation for cross-view gait recognition.
\newblock {\em IPSJ Transactions on Computer Vision and Applications}, 10(1):4,
  2018.

\bibitem{Wen2016A}
Y.~Wen, K.~Zhang, Z.~Li, and Y.~Qiao.
\newblock A discriminative feature learning approach for deep face recognition.
\newblock In {\em European conference on computer vision}, pages 499--515,
  2016.

\bibitem{Wu2017A}
Z.~Wu, Y.~Huang, L.~Wang, X.~Wang, and T.~Tan.
\newblock A comprehensive study on cross-view gait based human identification
  with deep cnns.
\newblock {\em IEEE Transactions on Pattern Analysis \& Machine Intelligence},
  39(2):209--226, 2017.

\bibitem{Yu2017GaitGAN}
S.~Yu, H.~Chen, E.~B.~G. Reyes, and N.~Poh.
\newblock {GaitGAN}: Invariant gait feature extraction using generative
  adversarial networks.
\newblock In {\em Proceedings of the IEEE Conference on Computer Vision and
  Pattern Recognition Workshops}, pages 30--37, 2017.

\bibitem{yu2019gaitganv2}
S.~Yu, R.~Liao, W.~An, H.~Chen, E.~B.~G. Reyes, Y.~Huang, and N.~Poh.
\newblock {GaitGANv2}: Invariant gait feature extraction using generative
  adversarial networks.
\newblock {\em Pattern recognition}, 87:179--189, 2019.

\bibitem{yu2006framework}
S.~Yu, D.~Tan, and T.~Tan.
\newblock A framework for evaluating the effect of view angle, clothing and
  carrying condition on gait recognition.
\newblock In {\em the 18th International Conference on Pattern Recognition},
  pages 441--444, 2006.

\bibitem{Yu2017View}
S.~Yu, Q.~Wang, L.~Shen, and Y.~Huang.
\newblock View invariant gait recognition using only one uniform model.
\newblock In {\em International Conference on Pattern Recognition}, pages
  889--894, 2017.

\bibitem{Zheng2011Robust}
S.~Zheng, J.~Zhang, K.~Huang, R.~He, and T.~Tan.
\newblock Robust view transformation model for gait recognition.
\newblock In {\em IEEE International Conference on Image Processing}, pages
  2073--2076, 2011.

\bibitem{zheng:rpca-vtm}
S.~Zheng, J.~Zhang, K.~Huang, R.~He, and T.~Tan.
\newblock Robust view transformation model for gait recognition.
\newblock In {\em ICIP}, pages 2073--2076, 2011.

\end{thebibliography}
}

\end{document}